\pdfoutput=1

\documentclass[11pt]{article}

\usepackage{acl}

\usepackage{times}
\usepackage{latexsym}

\usepackage[T1]{fontenc}

\usepackage[utf8]{inputenc}
\usepackage{amsmath}

\usepackage{microtype}

\usepackage{inconsolata}

\usepackage{graphicx}
\usepackage{amssymb}
\usepackage{wasysym}
\usepackage{xcolor}
\usepackage{colortbl}
\title{LLM$\times$MapReduce-V3: Enabling Interactive In-Depth Survey Generation through a MCP-Driven Hierarchically Modular Agent System}

\author{
Yu Chao$^1$\thanks{Equal contribution.} \
  Siyu Lin$^2$\footnotemark[1] \
  Xiaorong Wang$^3$ \
  Zhu Zhang$^{1}$ \
  Zihan Zhou$^{3}$ \\
  \textbf{Haoyu Wang}$^{4}$ \
  \textbf{Shuo Wang}$^1$\thanks{Corresponding authors.} \
  \textbf{Jie Zhou}$^3$ \
  \textbf{Zhiyuan Liu}$^{1}$\footnotemark[2] \
  \textbf{Maosong Sun}$^{1}$\footnotemark[2] \\
  $^{1}$Dept. of Comp. Sci. \& Tech., Institute for AI, BNRist Center, Tsinghua University \\
  $^2$Peking University \quad
  $^3$Modelbest Inc. \quad
  $^4$Nanyang Technological University \\
}

\definecolor{highlight}{RGB}{200, 0, 0}

\begin{document}
\maketitle
\begin{abstract}
We introduce {\bf LLM$\times$MapReduce-V3}, a hierarchically modular agent system designed for long-form survey generation. Building on the prior work, LLM$\times$MapReduce-V2, this version incorporates a multi-agent architecture where individual functional components, such as skeleton initialization, digest construction, and skeleton refinement, are implemented as independent model-context-protocol (MCP) servers. These atomic servers can be aggregated into higher-level servers, creating a hierarchically structured system. A high-level planner agent dynamically orchestrates the workflow by selecting appropriate modules based on their MCP tool descriptions and the execution history. This modular decomposition facilitates human-in-the-loop intervention, affording users greater control and customization over the research process. Through a multi-turn interaction, the system precisely captures the intended research perspectives to generate a comprehensive skeleton, which is then developed into an in-depth survey. Human evaluations demonstrate that our system demonstrates strong performance in both content depth and length, highlighting the strength of MCP-based modular planning.\footnote{The code is publicly available at \url{https://github.com/thunlp/LLMxMapReduce}.}
\end{abstract}

\section{Introduction}

Generating high-quality long-form survey articles poses significant challenges to AI Agent systems~\cite{hellobench2024,zhang2025mastering,wang2024factualitylargelanguagemodels}.  First, the system must effectively aggregate and synthesize information from a large collection of reference materials~\cite{nakano2021webgpt,asai2023self,huang2024researchagent}, which span various subdomains and perspectives. Second, it requires the ability to construct a coherent and comprehensive skeleton that guides the organization of content at a global level~\cite{wang2024autosurvey,wen2025interactive}. Third, as the upstream component, reference exploitation exerts a significant influence~\cite{gptresearcher2024}, thus many AI-based survey generation tools have begun to focus on improving the quality of search results before the writing stage begins. Addressing these challenges demands controllable mechanisms for information acquisition, structural planning, and content generation.

To address these challenges, we propose \textbf{LLM$\times$MapReduce-V3}, an interactive and self-organized modular agents system for long-form survey generation. Our system builds upon the design of LLM$\times$MapReduce-V2~\cite{wang2025llmtimesmapreducev2entropydrivenconvolutionaltesttime}, but extends it in several important ways to enable modularity, adaptability, and dynamic planning~\cite{qiu2025alitageneralistagentenabling,qiu2025agentdistilltrainingfreeagentdistillation}. At the core of our approach is the use of the model context protocol (MCP), a standardized function-calling mechanism that allows tools and modules to be composed as independent MCP servers~\cite{raseed2025mcp,unleash2025mcp}. We proposes a multi-stage workflow of document digestion, skeleton construction and refinement, and survey writing. Building upon this foundation, we re-architect the system into a multi-agent paradigm, wherein core functionalities are decomposed across specialized agents, and the algorithmic workflow is encapsulated as a suite of tools in MCP servers~\cite{anthropic2024mcp} to facilitate agent-level invocation and coordination.

A key improvement in our system lies in pipeline flexibility with an agent planner~\cite{crewai2024,ijcai2024p3}. Rather than following a fixed control flow, the agent receives a set of available MCP tools along with previous outputs, and dynamically selects the next module to invoke. This enables non-linear, adaptive workflows tailored to the specific needs of each writing task. To support user intent alignment, we also introduce \textbf{human-in-the-loop} interaction. Users begin by providing a target topic and basic writing instructions. The system then engages in a multi-turn dialogue to identify the user’s preferred and fine-grained research perspectives.

To thoroughly examine the performance of LLM$\times$MapReduce-V3, we conduct a human evaluation to assess system effectiveness. Domain experts compare the output of our system with those from other popular deep research systems across multiple topics. The results suggest that our system generates more informative outlines and higher-quality, more in-depth survey articles. Our main contributions are summarized as follows:
\begin{itemize}
 \item \textbf{Methodological Innovation}: We propose the first MCP-based
modular agent system for academic survey generation, enabling unprecedented
customization and institutional integration while maintaining survey quality
standards.
 \item \textbf{Architectural Advancement}: We propose a dynamic, LLM-driven planner that supports multi-stage module orchestration for adaptive, non-static workflows.
 \item \textbf{User-Centric Design}: We design a human-in-the-loop
interaction framework that ensures generated surveys align with user
expertise and research perspectives, bridging the gap between automation and
scholarly rigor.
\end{itemize}

\section{Related Work}

\subsection{AI-Powered Automated Research}

Early automated research systems focused on web-based information retrieval, with WebGPT~\cite{nakano2021webgpt} and Self-RAG~\cite{asai2023self} pioneering LLM-based web browsing through human feedback or self-reflection mechanisms for adaptive retrieval and generation. Recent advances include sophisticated autonomous agents like GPT-Researcher~\cite{gptresearcher2024} with dual-agent architectures, ResearchAgent~\cite{huang2024researchagent} for iterative research idea generation through multi-agents collaboration, and products including Perplexity Deep Research~\cite{perplexity2024} and ChatGPT Deep Research~\cite{openai2024chatgpt}. While some closed-source deep research systems can produce valuable results, we contend that flexible user involvement is critical for a truly practical system. We therefore propose an open-source, hierarchically modular system specifically designed to facilitate and support human intervention throughout the research process.

\subsection{Survey Generation}

LLM-driven survey generation has emerged with AutoSurvey~\cite{wang2024autosurvey}, which introduced a four-stage methodology that addresses context limitations and evaluation challenges, while InteractiveSurvey~\cite{wen2025interactive} advances through personalized, interactive generation with continuous user customization of reference categorization and content synthesis. SurveyX~\cite{liang2025surveyx}, on the other hand, focuses on extracting topics and content by pre-organizing the literature into the form of an attribute tree. Despite these advances, existing systems often lock users into a rigid, "all-or-nothing" paradigm. Stemming from inflexible integration approaches, they lack the necessary interfaces for iterative refinement and specialized customization. In response, we adopt MCP to modularize our algorithmic workflow, enabling the agents to iteratively select and invoke appropriate tools for survey generation.

\subsection{Model Context Protocol and Modular Self-Organized Agent System}

MCP~\cite{anthropic2024mcp} establishes open standards for connecting AI assistants to diverse tool sources through a unified client-server architecture. Recent applications demonstrate its potential for multi-agents intelligence~\cite{raseed2025mcp} and scalable systems~\cite{unleash2025mcp}. Alita~\cite{qiu2025alitageneralistagentenabling} leverages MCP to autonomously construct and reuse external capabilities through task-related protocols, achieving scalable agentic reasoning with minimal pre-definition. AgentDistill~\cite{qiu2025agentdistilltrainingfreeagentdistillation} enables training-free knowledge transfer via reusable MCP boxes, allowing smaller agents to achieve performance comparable to large LLM systems. These MCP-based self-evolution approaches demonstrate the its potential for adaptive agent systems.
Our work first introduces the MCP-based hierarchically modular survey generation system, combining protocol standardization with specialized academic optimizations to enable unprecedented user customization and institutional integration while maintaining the quality of survey paper generation.

\section{Hierarchically Modular Agent System}

\begin{figure*}
\centering
\footnotesize
  \includegraphics[width=0.99\linewidth]{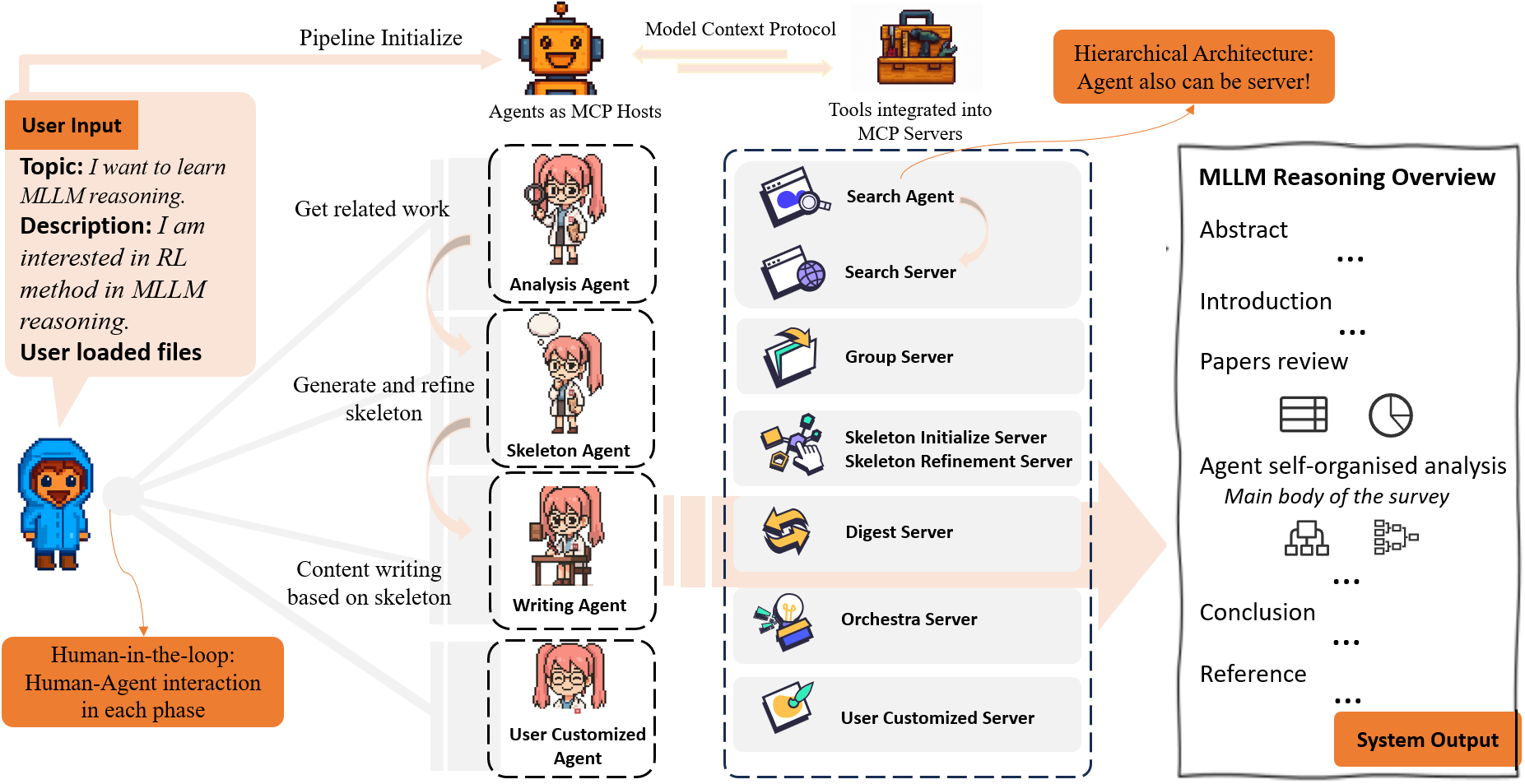} 
  \caption {Our agent-server ecosystem pipeline. Users begin by specifying a topic, optionally adding detailed descriptions or uploading documents. The Analysis Agent interprets user intent and coordinates with the Search Agent to retrieve and organize relevant literature. The Skeleton Agent then generates and refines an outline, which is used by the Writing Agent to complete the paper.}
  
\end{figure*}

Our system employs a multi-agent paradigm where specialized agents handle distinct phases of survey generation, each independently connecting to both algorithm-internal and user-provided MCP servers.

\subsection{System Design}

Let $\mathcal{A} = \{A_1, A_2, A_3\}$ denote the set of specialized agents, where $A_1$, $A_2$, $A_3$ denote the Analysis Agent, Skeleton Agent, and the Wirting Agent, respectively. The MCP server ecosystem can be represented as $\mathcal{S} = \{S_1, S_2, S_3, S_4, S_5, S_6, S^*\}$, including the Search Server, Group Server, Skeleton Initialization Server, Digest Server, Skeleton Refinement Server, Orchestra Server, and user customized servers (i.e., $S^*$). 

At each tool-calling round, the connection between the Agent and the Server is defined as $\mathcal{E}$, which is determined based on the previous output and the Agent's current plan.
$$ \mathcal{E} = \mathrm{MCP}(A_i(\mathrm{output}_{i-1}, \mathrm{plan}), \phi(\mathcal{A}_i))$$
The agent-server mapping is defined as:
$$\phi: \mathcal{A} \rightarrow 2^{\mathcal{S}}$$
where $\phi(A_i)$ specifies the server subset accessible to agent $A_i$.

The system architecture forms a directed graph $G = (\mathcal{A}, \mathcal{E}, \mathcal{S})$ where $\mathrm{MCP}(\mathcal{A}_i,\mathcal{S}_i)$  represents agents communicate through standardized MCP interfaces. 

Each server $S_i \in \mathcal{S}$ exposes a tool collection $\mathcal{T}(S_i) = \{t_{i,1}, t_{i,2}, ..., t_{i,k_i}\}$ through standardized MCP protocols. Tool invocation is formalized as:
$$\text{invoke}: \mathcal{A} \times \mathcal{T} \times \mathcal{I} \rightarrow \mathcal{O}$$
where $\mathcal{I}$ and $\mathcal{O}$ represent input and output spaces respectively.



\subsubsection{Analysis Agent}

The Analysis Agent orchestrates the initial literature processing and manages interaction with the Search Agent. Given a user-defined research topic and initial description, it first explores users' unwritten needs by multi-turn dialogue with users, thus enhance both retrieval width and depth. 

It then invokes the search agent for references acquisition and integrates materials uploaded by the user. Subsequently, the agent applies clustering algorithms from the group server to group the references by thematic relevance and methodological similarity, constructing a structured representation of the retrieve result. 
Finally, the agent initializes a tree graph to store the grouped references, and all downstream data structures are organized by this tree.

\subsubsection{Skeleton Agent}
The Skeleton Agent is responsible for constructing a globally organized and content-aware skeleton, serving as the structural backbone of the generated survey article. It operates under the coordination of an orchestration module and proceeds through three main stages: {\em Skeleton Initialization}, {\em Digest Construction}, and {\em Skeleton Refinement}.

\subsubsection{Writing Agent} 

The Writing Agent executes the final content synthesis, transforming the refined skeleton into coherent survey sections. This agent integrates literature digests, maintains citation consistency, and ensures academic writing standards while preserving the logical flow established by the skeleton structure.

To enhance generation quality, users may customize the writing agent to adapt to varying formatting specifications, support figure generation, and address other individualized requirements. Here we implement a dedicated figure server to support the generation of Mermaid-style diagrams.

\subsection{External and Replaceable Agents}

\subsubsection{Search Agent} 

The Search Agent provides literature retrieval capabilities through the interface. This agent can be seamlessly replaced with alternative search implementations, including domain-specific databases, institutional repositories, or specialized academic search engines. 

In parallel, the agent is also capable of acting as an MCP server. In our implementation, the Search Agent is supported by a dedicated search server, which integrates four core tools: query generation, web retrieval, crawling, and similarity analysis. This high-level server operates under the coordination of the Analysis Agent.

\subsubsection{Other Agents}

The system’s extensibility is reflected in its support for user-defined agents—such as academic critics, formatting agent, and domain-specific analyzers—enabling institutional customization and specialized workflows while preserving overall coherence through standardized interfaces.

\section{MCP Implementation Framework}

Our implementation leverages MCP's client-server architecture to encapsulate core functionalities as reusable, composable modules. Each functional component operates as an independent MCP server, exposing standardized tool interfaces for agent invocation.

\subsection{Native Server Construction}

We re-organize the procedure proposed in LLM$\times$MapReduce-V2~\cite{wang2025llmtimesmapreducev2entropydrivenconvolutionaltesttime} into several MCP servers.

\subsubsection{Group Server}
The Group Server serves as a content-based organizer that preprocesses the retrieved reference corpus. Before structural planning begins, this module clusters the documents into coherent topical groups. This pre-organization significantly reduces topic fragmentation and provides a more stable input for downstream outline construction. The grouping strategy is guided by both thematic relevance and methodological similarity.

\subsubsection{Orchestra Server}
The orchestration of the skeleton construction process is managed by the Orchestra Server, which acts as a lightweight planner based on an LLM backbone. At each stage, it takes the current intermediate outputs as input, along with formal descriptions of the available servers. Based on these inputs, it generates next-step instructions. This centralized coordination allows the system to adaptively guide the overall skeleton-building process, ensuring alignment between user intent, module capability, and evolving outline structure. The Orchestra Server thus serves as a dynamic planner and command generator that orchestrates multi-module collaboration across the pipeline.

\subsubsection{Skeleton Initialization Server}
Given the refined research angles and grouped references provided by the Analysis Agent, the Skeleton Initialization Server constructs a high-level section-wise outline. This initial structure serves as a coarse-grained scaffold, segmenting the article into major thematic areas that reflect distinct facets of the research topic. Each section is derived by mapping the user-specified focus onto prominent subfields within the references, ensuring both topical coverage and logical segmentation. 

\subsubsection{Digest Server}
The Digest Construction Server enhances the initial skeleton by generating content aware revision signals derived from the reference documents. It includes two steps: firstly, for each reference document, the system prompts an LLM to generate a brief summary along with suggestions for improving the current outline. These digests reflect how individual sources align or conflict with the existing skeleton. Secondly, the system aggregates all digests and suggestions, merges redundant feedback, and synthesizes a consolidated revision plan. This output serves as a high-level guide for improving the outline’s coverage, organization, and alignment with the source materials.

\subsubsection{Skeleton Refine Server}
Skeleton Refinement Server applies an iterative multi-layer convolution-inspired process to optimize the structure for coherence, consistency, and informativeness. This refinement mechanism operates both within sections (to enhance semantic alignment among digests) and across sections (to improve global representation and eliminate redundancy). This process simulates the multi-layer grouped convolution in Convolutional Neural Networks (CNN), which effectively integrates intra-group information and expand the contextual "receptive field" during information aggregation through a multi-layer iterative method.
When the references are effectively aggregated, Skeleton Refine Server produces an in-depth and fine-grained survey skeleton, which also determines the structural integrity of the subsequent survey writing.

\subsubsection{Iterative Refinement Through Multi-turn Tool-use}

The refinement process is conceptualized as a multi-turn, tool-based self-evolution framework, wherein the agent acts as a dynamic coordinator. This coordination is guided by intermediate outputs and, when available, user feedback. 

The Orchestra Server implements a planning function:
$$\pi: \mathcal{H} \times \mathcal{C} \rightarrow \mathcal{T}^*$$
where $\mathcal{H}$ denotes execution history, $\mathcal{C}$ represents current context, and $\mathcal{T}^*$ is the space of tool sequences.

At each decision point $t$, given skeleton state $x^{(t)} \in \mathcal{X}$ and history $h^{(t)} \in \mathcal{H}$, the agent selects action:
$$u^{(t)} = \pi(x^{(t)}, h^{(t)})$$

The refinement process operates through iterative state transitions:
$$x^{(t+1)} = f(x^{(t)}, u^{(t)})$$

Through this mechanism, the system progressively refines the survey skeleton and content analysis, enabling adaptive improvement across multiple iterations.

At each refinement iteration, skeleton agent consults the orchestra server to determine the optimal sequence of actions based on current skeleton state and available feedback. The Orchestra Server analyzes the current context and returns structured action plans which includes subsequent MCP server invocations, enabling adaptive workflow management based on real-time assessment of skeleton quality and user requirements.

Skeleton agent employs the Digest Server to generate targeted literature summaries for specific skeleton sections. 

These digests are then integrated into the skeleton structure through coordinated calls to theSkeleton Server, which maintains structural consistency while incorporating new content elements.

\subsection{Agent Integration and Replaceable Invocation}

Our orchestration mechanism treats each MCP server as a callable tool, associated with relevant metadata, which allows the central planner to make informed decisions regarding the sequencing of modules. 

Additionally, the system supports user-defined extensions, enabling the integration of custom MCP servers or external servers tailored to specific agents or tasks.

\begin{table*}
\centering
\footnotesize

\begin{minipage}{\textwidth}
\footnotesize
\centering
\textbf{Legend:} 
\textcolor{green}{$\checkmark$} = Full Support; 
\textcolor{orange}{$\sim$} = Limited Support; 
\textcolor{red}{$\times$} = No Support
\end{minipage}

\vspace{0.2cm}

\begin{tabular}{|l|c|c|c|c|c|c|}
\hline
\textbf{System} & \textbf{User} & \textbf{Modular} & \textbf{MCP} & \textbf{Custom} & \textbf{Survey} & \textbf{Open}\\
 & \textbf{Interaction} & \textbf{Design} & \textbf{Support} & \textbf{Tools} & \textbf{Focus} & \textbf{Source}\\
\hline
\multicolumn{6}{|c|}{\textbf{Commercial Platforms}} \\
\hline
Perplexity DR~\cite{perplexity2024} & \textcolor{green}{$\checkmark$} & \textcolor{red}{$\times$} & \textcolor{red}{$\times$} & \textcolor{red}{$\times$} & \textcolor{red}{$\times$} & \textcolor{red}{$\times$}\\
\rowcolor{gray!20}
OpenAI DR~\cite{openai2024chatgpt} & \textcolor{green}{$\checkmark$} & \textcolor{red}{$\times$} & \textcolor{red}{$\times$} & \textcolor{red}{$\times$} & \textcolor{red}{$\times$} & \textcolor{red}{$\times$} \\
Gemini DR~\cite{google_deep_research_2024} & \textcolor{green}{$\checkmark$} & \textcolor{red}{$\times$} & \textcolor{red}{$\times$} & \textcolor{red}{$\times$} & \textcolor{red}{$\times$} & \textcolor{red}{$\times$}\\
\rowcolor{gray!20}
Manus AI~\cite{manus_ai_2025} & \textcolor{green}{$\checkmark$} & \textcolor{red}{$\times$} & \textcolor{red}{$\times$} & \textcolor{red}{$\times$} & \textcolor{red}{$\times$} & \textcolor{red}{$\times$}\\
\hline
\multicolumn{6}{|c|}{\textbf{Deep Research Systems}} \\
\hline
WebGPT~\cite{nakano2021webgpt} & \textcolor{green}{$\checkmark$} & \textcolor{orange}{$\sim$} & \textcolor{red}{$\times$} & \textcolor{red}{$\times$} & \textcolor{red}{$\times$} & \textcolor{red}{$\times$} \\
\rowcolor{gray!20}
GPT-Researcher~\cite{gptresearcher2024} & \textcolor{orange}{$\sim$} & \textcolor{green}{$\checkmark$} & \textcolor{green}{$\checkmark$} & \textcolor{green}{$\checkmark$} & \textcolor{red}{$\times$} & \textcolor{green}{$\checkmark$}\\
ResearchAgent~\cite{huang2024researchagent} & \textcolor{green}{$\checkmark$} & \textcolor{green}{$\checkmark$} & \textcolor{red}{$\times$} & \textcolor{orange}{$\sim$} & \textcolor{red}{$\times$} & \textcolor{orange}{$\sim$}\\
\rowcolor{gray!20}
Self-RAG~\cite{asai2023self} & \textcolor{red}{$\times$} & \textcolor{green}{$\checkmark$} & \textcolor{red}{$\times$} & \textcolor{red}{$\times$} & \textcolor{red}{$\times$} & \textcolor{green}{\checkmark}\\
CoSearchAgent~\cite{cosearchagent2024} & \textcolor{orange}{$\sim$} & \textcolor{orange}{$\sim$} & \textcolor{red}{$\times$} & \textcolor{red}{$\times$} & \textcolor{red}{$\times$} & \textcolor{red}{$\times$}\\
\rowcolor{gray!20}
OpenResearcher~\cite{openresearcher2024} & \textcolor{red}{$\times$} & \textcolor{orange}{$\sim$} & \textcolor{red}{$\times$} & \textcolor{red}{$\times$} & \textcolor{red}{$\times$} & \textcolor{orange}{$\sim$}\\
Search-o1~\cite{searcho1_2025} & \textcolor{red}{$\times$} & \textcolor{orange}{$\sim$} & \textcolor{red}{$\times$} & \textcolor{red}{$\times$} & \textcolor{red}{$\times$} & \textcolor{green}{$\checkmark$}\\
\rowcolor{gray!20}
Agent-R1~\cite{agentr1_2025} & \textcolor{red}{$\times$} & \textcolor{orange}{$\sim$} & \textcolor{red}{$\times$} & \textcolor{red}{$\times$} & \textcolor{red}{$\times$} & \textcolor{green}{$\checkmark$}\\
\hline
\multicolumn{6}{|c|}{\textbf{Multi-Agent Frameworks}} \\
\hline
CrewAI~\cite{crewai2024} & \textcolor{orange}{$\sim$} & \textcolor{green}{$\checkmark$} & \textcolor{green}{$\checkmark$} & \textcolor{green}{$\checkmark$} & \textcolor{red}{$\times$} & \textcolor{green}{$\checkmark$}\\
\rowcolor{gray!20}
AutoAgent~\cite{ijcai2024p3} & \textcolor{red}{$\times$} & \textcolor{green}{$\checkmark$} & \textcolor{red}{$\times$} & \textcolor{orange}{$\sim$} & \textcolor{red}{$\times$} & \textcolor{green}{$\checkmark$}\\
Alita~\cite{qiu2025alitageneralistagentenabling} & \textcolor{red}{$\times$} & \textcolor{green}{$\checkmark$} & \textcolor{green}{$\checkmark$} & \textcolor{green}{$\checkmark$} & \textcolor{red}{$\times$} & \textcolor{green}{$\checkmark$}\\
\hline
\multicolumn{6}{|c|}{\textbf{Survey Generation Systems}} \\
\hline
AutoSurvey~\cite{wang2024autosurvey} & \textcolor{red}{$\times$} & \textcolor{red}{$\times$} & \textcolor{red}{$\times$} & \textcolor{red}{$\times$} & \textcolor{green}{$\checkmark$} & \textcolor{green}{$\checkmark$}\\
\rowcolor{gray!20}
InteractiveSurvey~\cite{wen2025interactive} & \textcolor{green}{$\checkmark$} & \textcolor{orange}{$\sim$} & \textcolor{red}{$\times$} & \textcolor{red}{$\times$} & \textcolor{green}{$\checkmark$} & \textcolor{green}{$\checkmark$}\\
SurveyX~\cite{liang2025surveyx} & \textcolor{orange}{$\sim$} & \textcolor{orange}{$\sim$} & \textcolor{red}{$\times$} & \textcolor{red}{$\times$} & \textcolor{green}{$\checkmark$} & \textcolor{green}{$\checkmark$}\\
\rowcolor{gray!20}
\textbf{LLM$\times$MapReduce-V3 (Ours)} & \textcolor{green}{$\checkmark$} & \textcolor{green}{$\checkmark$} & \textcolor{green}{$\checkmark$} & \textcolor{green}{$\checkmark$} & \textcolor{green}{$\checkmark$} & \textcolor{green}{\checkmark}\\
\hline
\end{tabular}
\caption{Comparison of Survey Generation and Deep Research Systems}
\label{tab:system_comparison}
\end{table*}

\section{Human-Agent Interaction}

Our system incorporates human feedback at key decision points to enhance survey quality and ensure alignment with user objectives. The interaction mechanism unfolds in structured phases that iteratively capture user input and refine system outputs.

\subsection{Consensus Achievement}

\textbf{Topic Consensus Phase} begins with user-defined topics and goals, followed by LLM-generated topic analysis and expansion. Through multi-turn dialogue, the system clarifies the research scope, identifies critical perspectives, and develops search strategies. This phase continues until a consensus is reached between the user and the AI on the survey's focus and coverage.

\subsection{Feedback Integration}

\paragraph{Outline Refinement Phase} presents the generated survey skeleton for user review and modification. Users may request structural adjustments, section reordering, or changes in content emphasis. The system processes this feedback via the skeleton server, ensuring coherence while accommodating user preferences.

\paragraph{Quality Assurance Integration} enables users to assess intermediate outputs at each stage, providing feedback that influences subsequent module executions. This ensures the final survey reflects both user expertise and the system's ability to synthesize comprehensive literature.


\section{Comparison and Evaluation}

As shown in Table~\ref{tab:system_comparison}, our LLM$\times$MapReduce-V3 system is uniquely equipped to meet the comprehensive demands of academic survey generation. While existing solutions excel in specific areas, commercial platforms in user interaction, research systems in modularity, and frameworks in MCP integration, none provide an all-encompassing solution.
Our system is the first to holistically integrate comprehensive user interaction, a modular design, MCP standardization, custom tool integration, and survey-specific optimizations. 

\begin{table}
  \centering
  \scalebox{1.0}{%
  \begin{tabular}{lrrr}
    \hline
    \textbf{System} & \textbf{Skeleton} & \textbf{Length} & \textbf{Quality}\\
    \hline
    Gemini DR     &     54.54\%    &   9.09\%    & 42.86\%\\
    Manus AI      &     27.27\%   &      9.09\%   & 0.00\%\\
    Our work      &     18.18\%    &   81.81\%   & 57.14\%\\\hline
  \end{tabular}
  }
  \caption{Human evaluation results. We recruited five human expert reviewers to evaluate articals generated by Gemini DeepResearch, Manus AI, and our system across eleven topics. Reviewers cast their votes based in three criteria: skeleton, length, and quality.}
  \label{tab:accents}
\end{table}

 The evaluation results in Table~\ref{tab:accents} indicate that Manus AI just generates preliminary outlines and contents lacks depth. Gemini DeepResearch provides superior depth, structural coherence, and fluency. Compared to other systems, ours provides broader coverage, particularly in literature reviews, and produces significantly longer contexts.

\section{Conclusion}


LLM$\times$MapReduce-V3 introduces a modular, MCP-based architecture that enables automated research assistance. By supporting open integration of customizable agents and servers, it overcomes the rigidity of traditional closed-agent systems. Through standardized interfaces, our system enables flexible composition of community-developed components to meet diverse research needs.
The proposed system also shows strong potential for broader application in knowledge-intensive tasks. With human-in-the-loop design, it achieves better alignment with human objectives.
Despite existing challenges, such as fixed workflows and complex data exchange, we advocate for open, adaptable agent ecosystems that evolve with advancing tools and demands.

\section*{Ethics Statement}
This work focuses on the development of an open-source, modular agent system for academic survey generation. Our system is intended to support and augment human researchers, not to replace them. We emphasize human-in-the-loop design to ensure transparency, user control, and alignment with scholarly standards. No sensitive personal data or copyrighted materials were used in model training or evaluation. All evaluation was conducted with the consent of expert annotators. We acknowledge the potential risks of misuse in automating academic writing and encourage responsible deployment with clear disclosure of AI assistance.

\section*{Acknowledgement}
This work is supported by the AI9Stars community. We also thank the anonymous reviewers for their insightful suggestions.

\bibliography{custom}

\appendix


\end{document}